\documentclass[letterpaper, 10 pt, conference]{ieeeconf}  

\IEEEoverridecommandlockouts                              

\usepackage{algorithm}
\usepackage{algorithmic}

\usepackage{amsmath}

\usepackage{graphicx}
\usepackage{listings}
\usepackage{multirow}
\usepackage[table,xcdraw]{xcolor}
\usepackage{adjustbox}
\usepackage{makecell}
\usepackage{graphicx}
\usepackage{subcaption}

\title{\LARGE \bf

Graph Representations for Reading Comprehension Analysis using Large Language Model and Eye-Tracking biomarker

}

\author{Yuhong Zhang$^{\dagger1}$,  Jialu Li$^{\dagger2}$, Shilai Yang$^{\dagger3}$, Yuchen Xu$^{1}$, Gert Cauwenberghs$^{1}$, Tzyy-Ping Jung$^{1}$
\thanks{$\dagger$ Equal Contribution}
\thanks{$^{1}$ Department of Bioengineering and Institute for Neural Computation, University of California, San Diego (UCSD), La Jolla, CA 92093, USA. {\tt\small yuz291@ucsd.edu, yux013@ucsd.edu, tpjung@ucsd.edu, gcauwenberghs@ucsd.edu}}%
\thanks{$^{2}$ Department of Electronic and Computer Engineering, Hong Kong University of Science and Technology, Hong Kong SAR. {\tt\small jlikr@connect.ust.hk}}%
\thanks{$^{3}$ School of Engineering, Brown University, Providence, RI 02912, USA. \texttt{shilai\_yang@brown.edu}}%
}

\begin{document}

\maketitle
\thispagestyle{empty}
\pagestyle{empty}

\begin{abstract}

Reading comprehension is a fundamental skill in human cognitive development. With the advancement of Large Language Models (LLMs), there is a growing need to compare how humans and LLMs understand language across different contexts and apply this understanding to functional tasks such as inference, emotion interpretation, and information retrieval. Our previous work used LLMs and human biomarkers to study the reading comprehension process. The results showed that the biomarkers corresponding to words with high and low relevance to the inference target, as labeled by the LLMs, exhibited distinct patterns, particularly when validated using eye-tracking data. However, focusing solely on individual words limited the depth of understanding, which made the conclusions somewhat simplistic despite their potential significance. This study used an LLM-based AI agent to group words from a reading passage into nodes and edges, forming a graph-based text representation based on semantic meaning and question-oriented prompts. We then compare the distribution of eye fixations on important nodes and edges. Our findings indicate that LLMs exhibit high consistency in language understanding at the level of graph topological structure. These results build on our previous findings and offer insights into effective human-AI co-learning strategies.\\


\end{abstract}


\section{INTRODUCTION}
Reading comprehension is a fundamental cognitive skill that remains a critical challenge in education and healthcare \cite{snow2002reading}. Globally, approximately 61\% of adolescents fail to achieve minimum reading comprehension proficiency by the end of lower secondary education \cite{unesco2017methodology,unesco2017reducing}. Studies have shown that attention deficits and low reading efficiency significantly impact students' ability to process and retain information effectively \cite{martinussen2005meta, mastropieri2003reading, vaughn2024teaching}. From a biomedical perspective, these challenges often stem from underlying neurological conditions, learning disabilities, or developmental disorders, making accurate assessment and intervention crucial for both educational and clinical applications. In the field of biomedical engineering, developing precise, quantitative methods for assessing cognitive function represents a significant challenge and opportunity for advancing healthcare delivery.

Recent LLMs advances have provided new insights into cognitive and human intelligence studies, particularly in reading comprehension tasks. This underscores the growing need to compare how humans and LLMs understand language across different contexts. Our previous work \cite{zhang2024integrating,zhang2024word} leveraged LLMs, particularly GPT-4, to label individual words in reading sentences, identifying whether they belong to groups of high relevance for question inference. Biomarkers from EEG and eye-tracking validated this approach, showing strong alignment with markers labeled with LLM. Although promising, this method has a limitation: sentence meaning often depends on phrases or word groups rather than individual words alone. This constraint makes it difficult to capture the broader semantic meaning of text and its relevance to comprehension tasks. Furthermore, word-level labeling fails to effectively characterize which parts of a sentence contain critical information that readers should focus on.  Readers often spend more time fixating on key phrases or sentence segments essential for understanding and answering questions rather than isolated words.

Researchers have investigated graph-based text representations to address the limitations of single-word representations. Graph-based representations depict words and phrases as nodes, with edges connecting the nodes representing their semantic relationships. Graph structures have long been used to represent sentence meanings \cite{sowa2014principles}, revealing the underlying logical flow of sentences and aligning closely with human cognitive processes \cite{lehmann1992semantic, brasoveanu2020semantics,brainsci11111424}. Unlike traditional bag-of-words (BoW) models, which assume term independence, graph-based text representations provide a structured approach to capturing linguistic relationships by modeling words or phrases as nodes and their syntactic or semantic connections as edges \cite{malliaros-vazirgiannis-2017-graph,rousseau-etal-2015-text}.

Beyond simple word co-occurrence, graph-theoretic techniques offer deeper insights into text analysis. Graph centrality measures—including degree centrality, PageRank, and betweenness centrality—identify key terms within a sentence, mirroring how humans prioritize information while reading. Additionally, k-core decomposition extracts dense subgraphs that highlight the most coherent and informative parts of a text. Similarly, Graph-of-Words (GoW) models capture concept interconnections more effectively,  reflecting the relationships readers form during text comprehension \cite{rousseau-etal-2015-text, rmus2022humans}.

We propose leveraging LLM-generated graph-based representations to advance reading comprehension analysis, integrating eye-tracking biomarkers, which have long been used in cognitive research. Prior studies have shown that eye-tracking data effectively predict reading comprehension performance \cite{10.3389/fpsyg.2022.1028824}.

This study examines various prompt heuristics and their corresponding knowledge graphs generated using LLMs. We analyze their topological structural features and compare them with LLM-generated labels to assess node importance, validating these findings with biomarkers. Our approach represents a novel contribution to biomedical engineering by establishing a framework that combines physiological measurements, artificial intelligence, and graph theory for cognitive assessment. The remainder of this paper is organized as follows: Section 2 introduces the Zurich Cognitive Language Processing (ZuCo) 1.0 datasets used for textual content and eye-tracking analysis. Section 3 presents our methodology for converting text sequences into graph structures and describes graph theory-based analysis methods. Section 4 discusses our results, focusing on graph measurement statistics and eye-gaze biomarkers. Finally, Section 5 summarizes our findings and outlines future research directions, particularly emphasizing potential clinical applications and biomedical implications for advancing healthcare technology and patient care.

\section{Dataset}
The ZuCo 1.0 dataset is a comprehensive multimodal resources that integrate high-density 128-channel EEG and eye-tracking data from 12 native English speakers engaged in various language comprehension and reading tasks \cite{hollenstein2018zuco}. It comprises 21,629 words, 1,107 sentences, and 154,173 fixations recorded over 4-6 hours of natural text reading. An EyeLink 1000 Plus eye tracker, operating at 500 Hz, tracked eye movements and pupil size.

This study fully leverages all tasks available in the ZuCo 1.0 datasets to analyze a range of eye-fixation features. High-resolution eye-tracking enables the examination of cognitive processes such as attention, language processing, and cognitive load, supporting research in psycholinguistics, cognitive neuroscience, and human-computer interaction\cite{hollenstein2018zuco}. 

The dataset includes three reading tasks:
\begin{itemize}
    \item Task 1 (Reading Task - Natural Sentences): Participants read grammatically correct and semantically meaningful English sentences in a natural reading setting.
    \item Task 2 (Reading Task - Random Words): Subjects read sequences of unconnected, randomly ordered English words, used to isolate low-level lexical processing from higher-order sentence integration.
    \item Task 3 (Reading Task - Baseline/Resting): Participants fixate on a central cross in the absence of any textual stimulus, providing a resting-state baseline for both eye and EEG signals.
\end{itemize}
\section{METHOD}
\subsection{Workflow Overview}

\begin{figure*}[ht]
    \centering
    \includegraphics[width=0.95\textwidth]{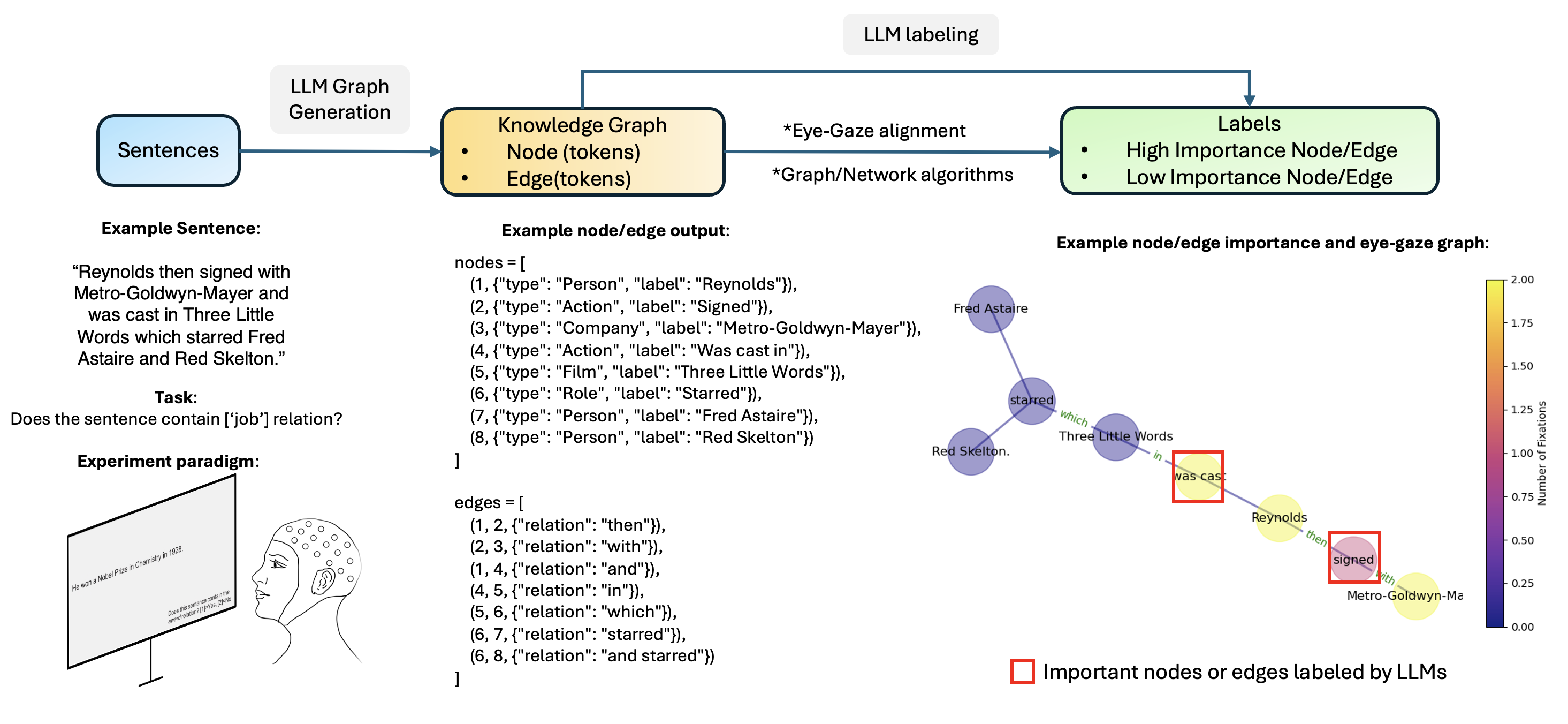} 
    \caption{Workflow for analyzing sentence semantics using LLM-generated graphs and eye-tracking data. Nodes represent entities (e.g., “Reynolds” and “Metro-Goldwyn-Mayer”), and edges capture relationships (e.g., “signed with”). The LLM highlights important nodes and edges marked with red boxes. This aids eye-tracking data and network analysis to identify critical job-related connections.}
    \label{fig:fig1}
\end{figure*}

Fig.~\ref{fig:fig1} illustrates the workflow. We begin by extracting sentences from three different tasks within the ZuCo 1.0 dataset. Based on semantic and logical transitions, we develop corresponding prompts to convert sentences into nodes and edges. Next, we use task-oriented reading questions to generate additional prompts that direct LLMs to label key nodes and edges, reflecting their internal reasoning process. Graph metrics, such as node centrality (degree, closeness, betweenness), are calculated to characterize graph properties and compared with the results provided by the AI agent. More importantly, we align human eye-tracking biomarkers to validate the computational results, specifically the number of fixations on nodes and edges.


\subsection{LLM and Knowledge Graph Generation}
 Extracting knowledge graphs (KGs) and identifying important nodes leverages LLMs through two carefully designed prompts and evaluation criteria. The following algorithm \ref{alg:alg1} outlines the methodology, which comprises two primary parts: (1) Knowledge Graph Generation and (2) Important Node Extraction.

\subsubsection{\textbf{Step 1:} Knowledge Graph Generation and Evaluation}
KGs are generated iteratively for each sentence in the ZuCo 1.0 dataset. To ensure the robustness of the extracted graphs, we performed multiple iterations for each sentence using Prompt 1-based querying; see the Appendix for Prompts 1 and 2. Each generated KG undergoes evaluation based on its fidelity to the input sentence, focusing on three key error metrics: omitted, extra, and mispronounced words. The process selects the most accurate graph, prioritizing those with the fewest missing words. Additional steps correct mispronounced words and remove extra words, thus ensuring that the final KG accurately represents the sentence with minimal error.

\subsubsection{\textbf{Step 2:} Important Node Extraction}
The second stage focuses on extracting important nodes from the generated KGs. Using Prompt 2, the LLM evaluates the importance of each node in the KG generated from Step 1 based on contextual and structural relevance. Nodes receive binary importance labels (0 or 1), which are then integrated into the final KG representation.  Including control questions and target words from the original ZuCo dataset, further refines node importance classification. 
\begin{algorithm}[htbp]
\caption{Knowledge Graph Construction and Important Node Extraction}
\renewcommand{\algorithmicrequire}{\textbf{Input:}}
\renewcommand{\algorithmicensure}{\textbf{Output:}}
\label{alg:alg1}
\begin{algorithmic}[1]
\REQUIRE 
  Dataset ZuCo 1.0, 
  Prompt $prompt \in \{1, 2\}$ for GPT-4o queries, 
  Loop iterations $loop\_time = 3$

\ENSURE 
  A knowledge graph per sentence $KG$ with each node importance assigned (0 or 1)
\vspace{2mm}
\FOR{each sentence $s_i$ in $dataset$}
    \STATE \textbf{Part 1: Knowledge Graph Generation}
    
    \FOR{$j = 1$ to $loop\_time$}
        \STATE $\mathit{llmOutput1} \gets \text{GPT-4o}(s_i, \text{prompt 1})$
        \STATE $KG_{ij} \gets \textit{parseLLMOutput}(\mathit{llmOutput1})$
        
        \STATE Compare $KG_{ij}$ with $s_i$ to obtain $error_{ij}$
        \STATE Append $(KG_{ij}, error_{ij})$ to the $trials_i$ of  $s_i$
    \ENDFOR

    \STATE  $KG_i$ is the one with the minimal missing words among $trials_i$

    \STATE Update  $KG_i$ by deleting extra and correcting mispronounced words 
    
\vspace{2mm}
    \STATE \textbf{Part 2: Important Node Extraction}

   \STATE Query GPT-4o for node importance assignment using $\bigl(\text{prompt 2},\, s_i,\, KG_i\bigr)$, 
       optionally including $\bigl(\text{control\_question},\, target\_word\bigr)$ if available, 
       to obtain $llmOutput2$.

    \STATE Parse $llmOutput2$ to assign importance for nodes

    \STATE Update nodes in $KG_i$ with importance assignment 
\ENDFOR

\end{algorithmic}
\end{algorithm}

\subsubsection{Prompting Heuristic}
As shown in Table \ref{llm_method_table_transformed}, zero-shot prompting and the Chain of Thought (CoT) technique enhance the accuracy and generalization of knowledge graph extraction by avoiding biases and promoting structured reasoning. Zero-shot prompting outperforms few-shot prompting, particularly in tasks requiring structural consistency, achieving higher accuracy and more consistent node importance. It avoids biases from example inputs, generalizes across diverse sentence structures, and reveals unique cyclical patterns in results. Additionally, CoT improves performance in complex tasks by guiding step-by-step reasoning, reducing errors, and ensuring consistent node importance assignments, highlighting the potential of structured reasoning in knowledge graph extraction. Comparison of various prompting heuristics based on different knowledge graph measurements using ChatGPT-4o. “Omit” refers to missing words, “Extra” refers to insertion errors, and “Misspelled” refers to words with spelling errors. “Tot.” represents the total error, calculated as the sum of these metrics, as indicated in Table \ref{llm_method_table_transformed}.

\subsection{Graph Semantics}

\subsubsection{Node Centrality criteria}

We model the text in the reading sentences as unweighted, directed graphs, $G = (V, E)$, where $V$ represents the set of nodes (e.g., words or phrases) and $E$ represents the set of directed edges (e.g., semantic or syntactic relations). In this framework, directed edges indicate specific dependencies, allowing us to more effectively infer relationships between different sentence components.  

We use several centrality metrics to identify important nodes in the network, including PageRank, Degree Centrality, Betweenness Centrality, and Closeness Centrality. Each metric evaluates node importance based on different structural properties of the graph. For instance, PageRank assesses the influence of a node based on its neighbors, while Degree Centrality quantifies the number of direct connections a node has. Table \ref{centrality_measure_table} provides a detailed overview of these metrics.

\begin{table}[htbp]
\centering
\renewcommand{\arraystretch}{1.5}
\caption{Comparison of various prompting heuristics based on different knowledge graph measurements}
\label{llm_method_table_transformed}
\begin{tabular}{l l l l l}
\hline
\textbf{Method}         & \textbf{Omit} & \textbf{Extra} & \textbf{Misspelled} & \textbf{Tot.} \\ \hline
Zero-Shot               & 0.567         & 0.020          & 0.0013            & 0.5883        \\ 
Few-Shot                & 0.720         & 0.030          & 0.002             & 0.7232        \\ 
COT Zero-Shot            & 0.520         & 0.021          & 0.0023            & 0.5433        \\ 
COT Few-Shot            & 0.525         & 0.021          & 0.0023            & 0.5483        \\ \hline
\end{tabular}
\end{table}

\begin{table}[htbp]
\centering
\renewcommand{\arraystretch}{1.2}
\caption{Definitions of Centrality Measures}
\label{centrality_measure_table}
\begin{tabular}{l l}
\hline
\textbf{Measure (Version)} & \textbf{Definition (short)} \\ \hline
PageRank (Unweighted)      & Importance on unweighted neighbors. \\ 
Degree Centrality          & Connections (degree) of a node. \\ 
Betweenness Centrality     & Frequency in shortest paths. \\ 
Closeness Centrality       & Proximity to other nodes. \\ \hline
\end{tabular}
\end{table}

\begin{table*}[t!]
\centering
\renewcommand{\arraystretch}{1.3}
\caption{Graph Statistics for Tasks 1-3 (ZuCo 1.0)}
\label{graph_stats_table}
\begin{tabular}{l | c c c c c}
\hline
\textbf{Task} & \textbf{Total Graphs} & \textbf{Disconnected Graphs} & \textbf{Disconnection (\%)} & \textbf{Avg Nodes} & \textbf{Avg Degree} \\ \hline
Task 1        & 600                   & 85                           & 14.167               & 3.862              & 1.508               \\
Task 2 (w/)   & 67                    & 5                            & 7.463                & 4.097              & 1.484               \\
Task 2 (w/o)  & 527                   & 45                           & 8.539                & 5.299              & 1.594               \\
Task 3        & 407                   & 25                           & 6.143                & 5.037              & 1.560               \\ \hline
\end{tabular}

\vspace{1em}

\begin{tabular}{l | c c c c c c}
\hline
\textbf{Task} & \textbf{Avg Path Len.} & \textbf{Avg Clust. Coef.} & \textbf{Avg Diam.} & \textbf{Avg Density} & \textbf{Avg Edges} & \textbf{Avg Graph Rank} \\ \hline
Task 1        & 1.518                  & 0.015                     & 2.577              & 0.614                & 3.012              & 3.103                    \\
Task 2 (w/)   & 1.562                  & 0.023                     & 2.452              & 0.548                & 3.145              & 2.952                    \\
Task 2 (w/o)  & 1.821                  & 0.015                     & 3.145              & 0.452                & 4.378              & 3.913                    \\
Task 3        & 1.733                  & 0.008                     & 2.853              & 0.468                & 4.089              & 3.497                    \\ \hline
\end{tabular}

\vspace{1em}

\begin{tabular}{l | c c c c c}
\hline
\textbf{Task} & \textbf{Avg Triangles} & \textbf{Star Graph (\%)} & \textbf{Cycle Graph (\%)} & \textbf{Path Graph (\%)} & \textbf{Complete Graph (\%)} \\ \hline
Task 1        & 0.023                  & 30.68                    & 13.981                   & 67.379                   & 17.67                         \\
Task 2 (w/)   & 0.048                  & 48.387                   & 3.226                    & 53.226                   & 9.677                         \\
Task 2 (w/o)  & 0.033                  & 29.461                   & 6.639                    & 41.494                   & 4.564                         \\
Task 3        & 0.021                  & 39.005                   & 5.236                    & 40.314                   & 4.45                          \\ \hline
\end{tabular}
\end{table*}

\subsubsection{ROC-AUC Calculation for Two Labeling System}
To compare node importance generated by LLMs and graph-theoretic metrics, we use Receiver Operating Characteristic - Area Under the Curve (ROC-AUC) as the evaluation metric. LLMs provide binary classifications (0 or 1) for node importance, while graph-theoretic metrics, such as unweighted PageRank, weighted PageRank, degree centrality, and betweenness centrality, generate continuous scores. Treating the LLM labels as ground truth, we compute the true positive rate (TPR) and false positive rate (FPR) across various thresholds for the graph-theoretic scores. The ROC curve is then plotted, and the area under the curve (AUC) quantifies the alignment between the two labeling systems. A higher AUC indicates stronger agreement between LLM predictions and centrality metrics. This approach establishes a framework for evaluating the consistency of node importance identification across both systems, with results further validated using eye-tracking biomarkers.

\subsection{Eye Fixation Graph Alignment}
In the ZuCo dataset, eye fixation refers to the stable gaze position maintained on a specific word or token during natural reading, with each fixation annotated by its number and position \cite{hollenstein2018zuco}. Instead of analyzing fixations at the word level, we aggregate the eye-gaze data into corresponding graph nodes or edges, following the abstraction method in \cite{zhang2024integrating}, enabling alignment with graph-based representations.
We then conduct statistical analysis to evaluate their significance. Specifically, we compare the average fixation number between important and non-important nodes to assess their significance.

\section{Result Analysis}

\subsection{Knowledge Graph Analysis}
This section examines the results of graph centrality measurements to determine the relative importance of nodes within the network. We then present the ROC curves and AUC for tasks in ZuCo 1.0, illustrating the alignment between node importance in each graph and the LLMs’ predictions.

\subsubsection{Graph and Centrality Measurements}

\begin{figure*}[htbp]
    \centering
   
        \begin{subfigure}[b]{0.15\textwidth}
        \includegraphics[width=\textwidth]{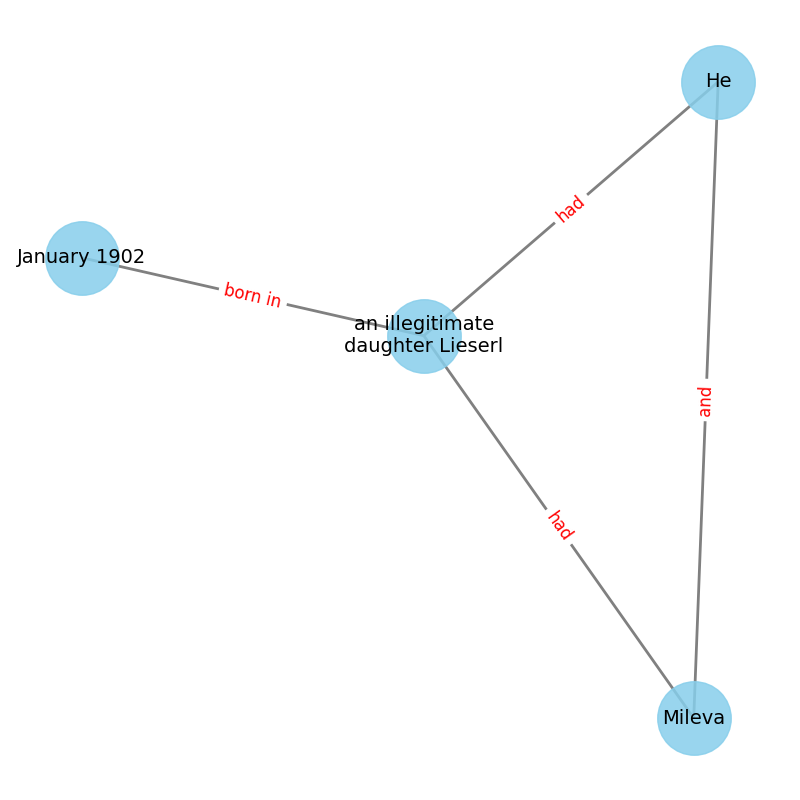}
       \caption{Triangle pattern forming a closed loop among three nodes.}
        \label{fig:pattern_triangle}
    \end{subfigure}
    \hfill
    \begin{subfigure}[b]{0.15\textwidth}
        \includegraphics[width=\textwidth]{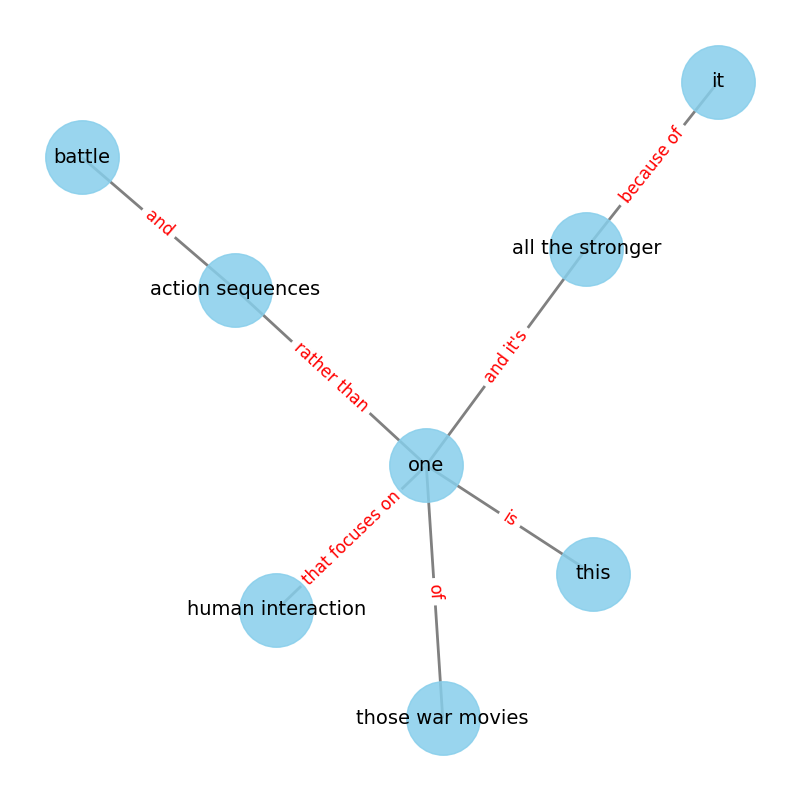}
        \caption{Star structure indicating central-hub organization.}
        \label{fig:pattern_star}
    \end{subfigure}
    \hfill
    \begin{subfigure}[b]{0.15\textwidth}
        \includegraphics[width=\textwidth]{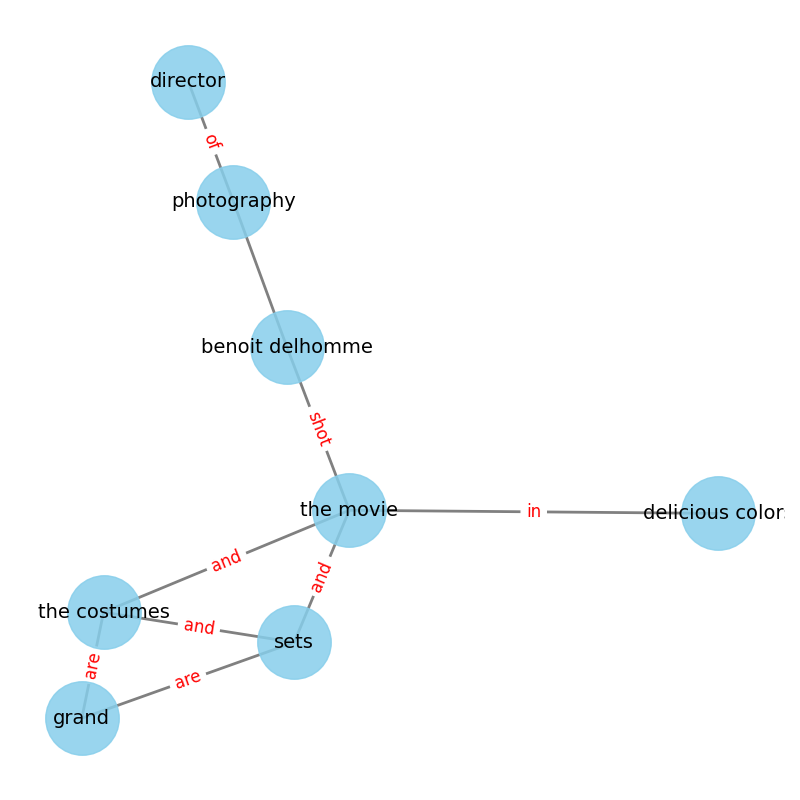}
        \caption{Cycle pattern representing closed relational loops.}
        \label{fig:pattern_cycle}
    \end{subfigure}
    \hfill
        \begin{subfigure}[b]{0.15\textwidth}
        \includegraphics[width=\textwidth]{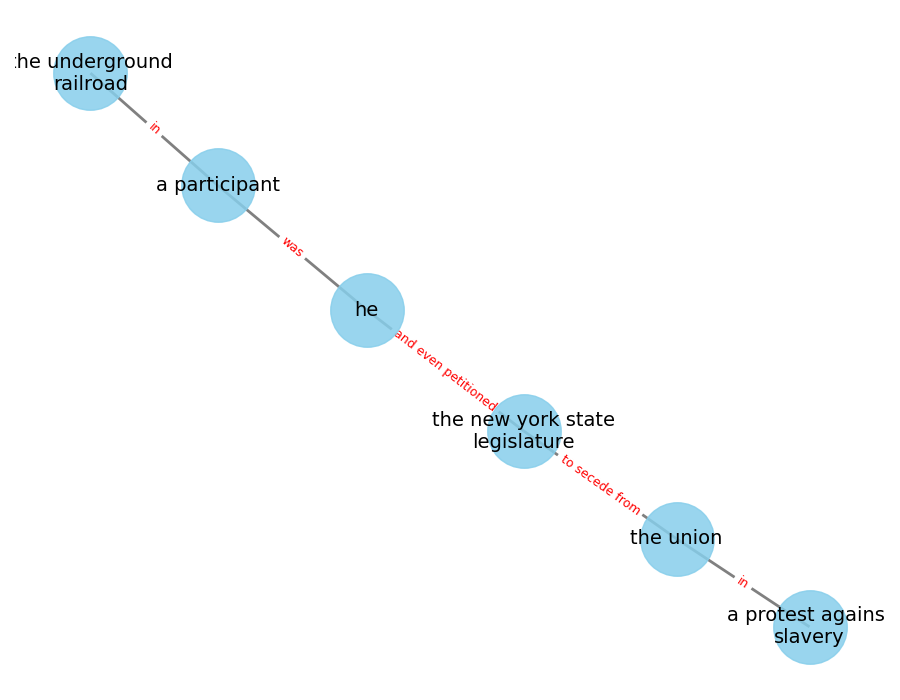}
        \caption{Path pattern with strong sequential dependencies.}
        \label{fig:pattern_path}
    \end{subfigure}
    \hfill 
    \begin{subfigure}[b]{0.15\textwidth}
        \includegraphics[width=\textwidth]{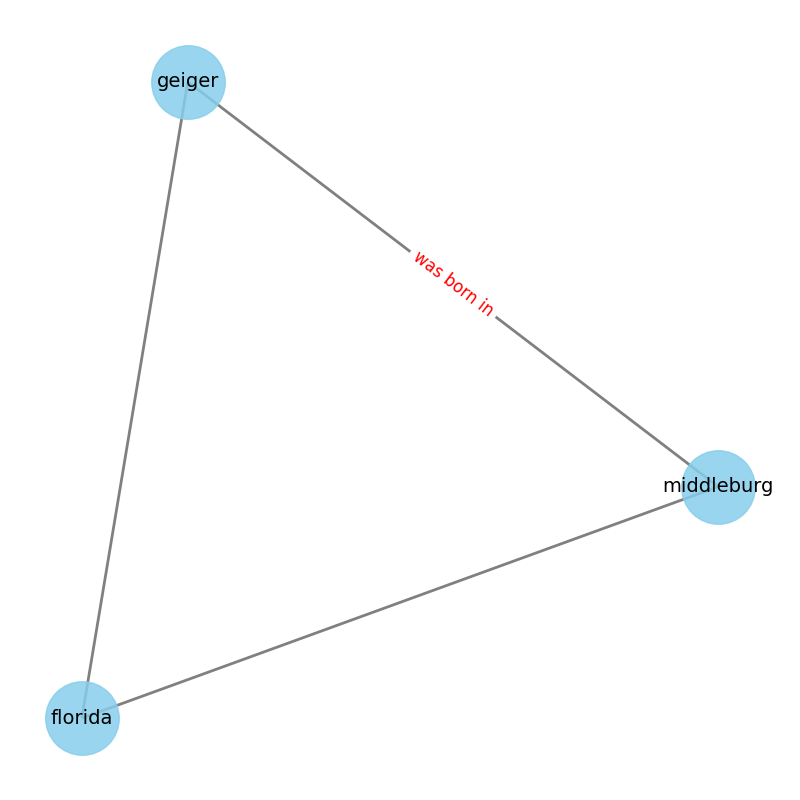}
        \caption{Complete graph with all-pair connectivity.}
        \label{fig:pattern_complete}
    \end{subfigure}
    \caption{Representative graph patterns extracted from ZuCo-1 dataset.}
    \label{fig:graph_patterns}
\end{figure*}

Table~\ref{graph_stats_table} summarizes key graph statistics across tasks 1-3, offering insights into structural variations that influence centrality-based analyses. Task 1 exhibits the highest percentage of disconnected graphs (14.167\%), indicating a more fragmented network structure than Task 3, which has the lowest disconnection rate (6.143\%). This suggests that centrality measures relying on global connectivity, such as Betweenness Centrality, may be less effective in Task 1 because of the lack of well-defined shortest paths. In contrast, Task 2 (without questions) features the highest average node count (5.299) and degree (1.594), indicating a denser graph where PageRank and Degree Centrality may better rank important nodes. The clustering coefficient is notably low across all tasks, with Task 3 exhibiting the lowest value (0.008), indicating sparsely connected graphs where localized clustering-based centrality measures may be less reliable. Additionally, the differences in average path length and network diameter across tasks indicate variations in how information propagates within the graphs, directly impacting how centrality-based methods assess node importance. Task 2 (without questions) also demonstrates the highest edge count (4.378) and triangle count (0.033), suggesting stronger connectivity and redundancy in graph structure. 

As illustrated in Figure~\ref{fig:graph_patterns}, the graph type distributions emphasize structural variations across tasks, with significant class overlap. Path graphs are prevalent across all tasks but are most dominant in Task 1 (67.379\%) and Task 2 (w/) (53.226\%), suggesting strong sequential dependencies. Star graphs are notably high in Task 2 (w/) (48.387\%) and Task 3 (39.005\%), indicating hierarchical relationships where central nodes connect to multiple others. Cycle graphs occur infrequently, peaking in Task 1 (13.981\%) and appearing least in Task 2 (w/) (3.226\%), indicating minimal cyclic dependencies. Complete graphs are rare across all tasks, with Task 1 (17.67\%) exhibiting the highest occurrence. The varied distribution of graph types suggests that Task 1 favors sequential and interconnected structures, Task 2 (w/) displays more hierarchical organization, Task 2 (w/o) shows moderate diversity in connectivity, and Task 3 balances multiple structures. These differences could impact the applicability of centrality measures, as hierarchical structures may favor Degree Centrality, sequential graphs benefit Path-based measures, and denser graphs support PageRank-based rankings.

\subsubsection{ROC-AUC Analysis}

\begin{figure}[ht]
    \centering
    \includegraphics[width=0.5\textwidth]{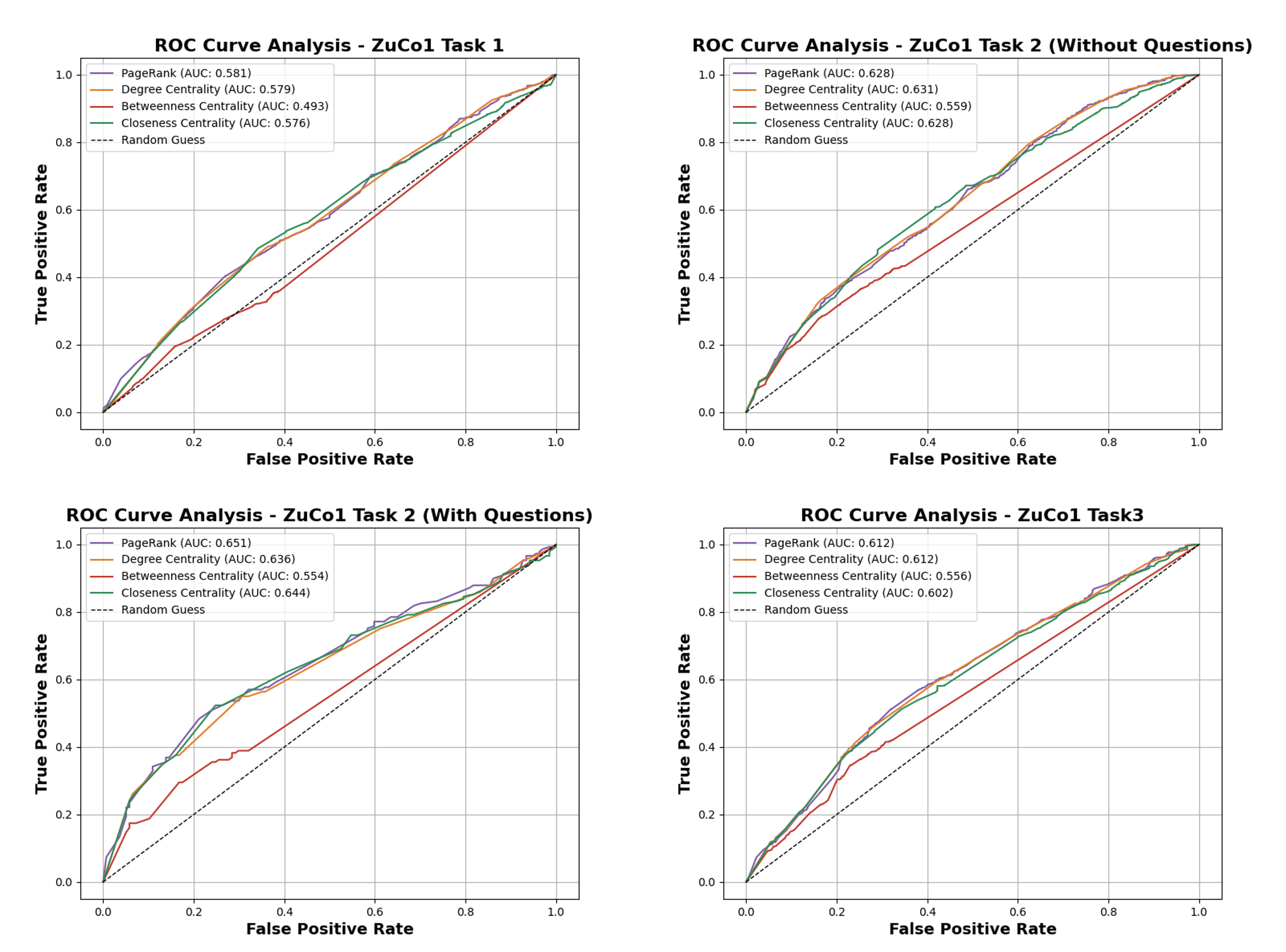} 
    \caption{ROC Curve Analysis for Tasks 1-3 in ZuCo 1.0. Each plot compares the AUC performance of PageRank, Degree Centrality, Betweenness Centrality, and Closeness Centrality metrics. PageRank consistently achieves the highest AUC across tasks, particularly for Task 2 (with questions), surpassing all other metrics. The results highlight the influence of network structure on the effectiveness of centrality-based node importance assessments.}
    \label{fig:roc_curves}
\end{figure}

Through testing and evaluation, selecting nodes within the top 50\% of PageRank values produced the best results for identifying significant nodes, as evidenced by the ROC-AUC analysis shown in Fig.~\ref{fig:roc_curves}.

The ROC curves illustrate the comparative performance of the centrality metrics across tasks, with the ground truth derived from LLM-generated labels for important node extraction. For Task 1, PageRank achieves the highest AUC (0.581), followed closely by Degree Centrality (0.579), while Betweenness Centrality performs significantly worse with an AUC of 0.493. For Task 2 (without questions), both PageRank and Degree Centrality show strong performance with AUCs of 0.628 and 0.631, respectively, while Betweenness Centrality remains lower at 0.559. Task 2 (with questions) highlights PageRank as the most effective metric, achieving an AUC of 0.651, followed by Closeness Centrality with an AUC of 0.644. In Task 3, PageRank and Degree Centrality achieve an AUC of 0.612, while Closeness Centrality falls slightly behind at 0.602. Treating the LLM-generated importance labels as ground truth, the PageRank results demonstrate relative consistency between generative AI models and graph-based analysis, supporting the validity of the generated results.

\subsection{Eye-fixation and Graph Alignment}

\begin{figure}[ht]
    \centering
    \includegraphics[width=0.45\textwidth]{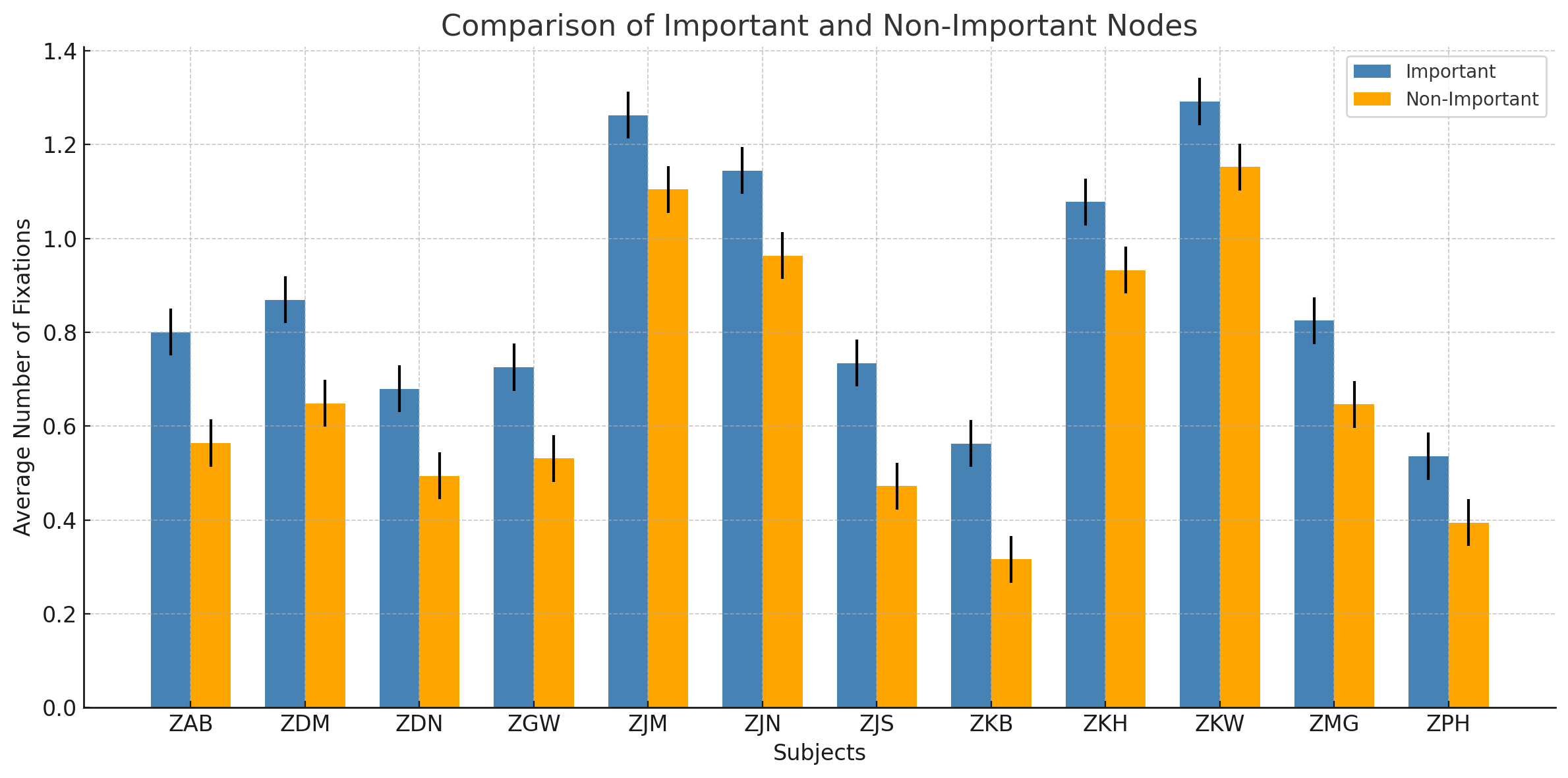} 
    \caption{Comparison of Important and Non-Important Nodes based on the number of average eye-fixation alignments in ZuCo 1.0. This graph shows the average number of fixations for important and non-important nodes across different subjects, with error bars representing the standard error (SE). Important nodes consistently exhibit higher fixation averages.}
    \label{fig:eye_fixation_graph}
\end{figure}




We analyzed the average fixation data for nodes classified as important and non-important across 12 subjects to quantify the alignment between eye fixations and graph-based node importance. 
As shown in Fig.~\ref{fig:eye_fixation_graph}, the average fixation for important nodes was consistently higher than that for non-important nodes across all subjects. Specifically, the mean fixation value for important nodes ranged from 0.562 (subject ZKB) to 1.292 (subject ZKW), while non-important nodes ranged from 0.316 (subject ZKB) to 1.152 (subject ZKW). 

Subjects such as ZKW and ZJM demonstrated the largest fixation averages for important nodes (1.292 and 1.263, respectively), showing substantial differences from non-important nodes (1.152 and 1.105, respectively). In contrast, subjects like ZAB and ZPH exhibited smaller fixation differences between important and non-important nodes, with differences of 0.236 and 0.141, respectively. These variations may reflect differences in individual reading strategies or cognitive focus.
Error bars in the figure represent the standard error (SE). The SE values for important nodes ranged from 0.035 (ZPH) to 0.064 (ZKW), while non-important nodes had SE values ranging from 0.030 (ZJS) to 0.061 (ZKW). The relatively low SE across subjects indicates a consistent pattern of fixation differences between the two node categories.
Quantitatively, the mean difference in fixation between important and non-important nodes across all subjects was 0.291, with a standard deviation of 0.111. 
This consistent difference highlights the strong alignment between LLM-based node importance labeling—established through consensus across  LLM outputs—and human cognitive attention, as measured by eye fixations. This serves as additional behavioral validation of the LLM-derived ground truth used in ROC-AUC analysis.

\section{CONCLUSIONS}

This study explored a novel graph-based text representation method using LLMs alongside eye-gaze biomarkers for reading comprehension tasks. This approach addresses the limitations of traditional BoW or single-word representations, which treat words as independent entities. This method uses graphs with nodes and edges to capture relationships between phrases in the sentence, such as co-occurrences within a text window, syntactic relationships, and semantic relationships.
We validated the graph-based results through mathematical intuition using graph and network measurements and human eye-gaze alignments. The ROC-AUC analysis and fixation statistics results confirmed the effectiveness of the LLM-generated graph representations, highlighting the generative model's ability to align with human cognitive processes. 

This study used a limited and controlled reading corpus. Future work must test the proposed pipeline in more diverse and real-world reading environments with varied 
materials and content. 
Additionally, integrating other biomarkers, such as EEG data, could provide more comprehensive insights into cognitive processing and potentially lead to more accurate diagnostic tools. This multi-modal approach could be particularly valuable in understanding and treating complex neurological conditions affecting reading comprehension.

From a biomedical engineering standpoint, our work contributes to the growing field of AI-assisted healthcare by demonstrating how machine learning and physiological measurements can be combined to create more effective diagnostic and therapeutic tools.



\bibliographystyle{ieeetr}
\bibliography{references}

 
 
 \section*{APPENDIX}
\subsection{Prompt 1: Building a Knowledge Graph from a Sentence}
\begin{quote}
\small
\ttfamily
Your task is to build a structured and accurate knowledge graph that captures the semantic meaning of the sentence. \\
Let's think step by step. \\
First, identify all minimal semantic units of the main entity, including articles like "the" and possessive pronouns like "his" in the node phrase. \\
Then, all the words in the original sentence except the node contents become the relations (edges). \\
Last, format the output into the format below: \\
\texttt{<nodes>} \\
\texttt{(node\_number, \{"type": the\_type\_of\_label, "label": node\_content\}),} \\
\texttt{</nodes>} \\
\texttt{<edges>} \\
\texttt{(starting\_node\_number, ending\_node\_number, \{"relation": phrase\_content\}),} \\
\texttt{</edges>} \\
The node number should start from 1. \\
Now you have to process this sentence: \texttt{{sentence}}
\end{quote}
\subsection{Prompt 2: Extracting Important Nodes from the Knowledge Graph}
\begin{quote}
\small
    \ttfamily
    Your task is to extract the important nodes from the given knowledge graph. \\
    Let's think step by step. \\
    You should determine which nodes are important based on the provided target words and the nodes' relevance to the core message of the sentence. \\
    Use this approach to identify the important nodes based on the following inputs: \\
    1. Input 1: the sentence: \texttt{{sentence}} \\
    2. Input 2: nodes in the knowledge graph for the sentence: \texttt{{nodes}} \\
    3. Input 3: the target word: \texttt{{target\_words}} \\
    Ensure that the output follows the format below: \\
    \texttt{<nodes>} \\
    \texttt{(node\_number, \{"type": the\_type\_of\_label, "label": node\_content\}),} \\
    \texttt{</nodes>}
\end{quote}

\end{document}